\documentclass{article}
\usepackage{spconf,amsmath,graphicx}
\usepackage{times}
\usepackage{url}
\usepackage{latexsym}
\usepackage{paralist}
\usepackage{multirow}
\usepackage{amssymb,verbatim}

\title{Dynamically Context-Sensitive Time-Decay Attention\\ for Dialogue Modeling}
\name{Shang-Yu Su\quad Pei-Chieh Yuan\quad Yun-Nung Chen}
\address{National Taiwan University, Taipei, Taiwan\\
\texttt{\small f05921117@ntu.edu.tw\quad b03901134@ntu.edu.tw\quad y.v.chen@ieee.org}}
\begin{document}
\ninept
\maketitle
\begin{abstract}
Spoken language understanding (SLU) is an essential component in conversational systems.
Considering that contexts provide informative cues for better understanding, history can be leveraged for contextual SLU.
However, most prior work only paid attention to the related content in history utterances and ignored the temporal information.
In dialogues, it is intuitive that the most recent utterances are more important than the least recent ones, and time-aware attention should be in a decaying manner. 
Therefore, this paper allows the model to automatically learn a time-decay attention function where the attentional weights can be dynamically decided based on the content of each role's contexts, which effectively integrates both content-aware and time-aware perspectives and demonstrates remarkable flexibility to complex dialogue contexts. 
The experiments on the benchmark Dialogue State Tracking Challenge (DSTC4) dataset show that the proposed dynamically context-sensitive time-decay attention mechanisms significantly improve the state-of-the-art model for contextual understanding performance.
% \footnote{The code will be released once accepted.}.
\end{abstract}

\section{Introduction}
\label{sec:intro}

Spoken dialogue systems that can help users solve complex tasks such as booking a movie ticket have become an emerging research topic in artificial intelligence and natural language processing areas. 
With a well-designed dialogue system as an intelligent personal assistant, people can accomplish certain tasks more easily via natural language interactions. 
%Today, there are several virtual intelligent assistants on the market, such as Apple's Siri, Google's Home, Microsoft's Cortana, and Amazon's Echo. 
The recent advance of deep learning has inspired many applications of neural dialogue systems~\cite{wen2017network,bordes2017learning,dhingra2017towards,li2017end}. %, su2018natural, su2018discriminative, su2018investigating}.

A key component of a dialogue system is a spoken language understanding (SLU) module---parsing user utterances into semantic frames that capture the core meaning~\cite{tur2011spoken}.
A typical pipeline of SLU is to first determine the domain given input utterances, and based on the domain to predict the intent and to fill the associated slots corresponding to a domain-specific semantic template~\cite{hakkani2016multi,chen2016knowledge,chen2016syntax,wang2016learning}.
%, where each utterance is treated independently~\cite{hakkani2016multi,chen2016knowledge,chen2016syntax,wang2016learning}.
% With the power of deep learning, there are emerging better approaches of SLU
However, the above work focused on single-turn interactions, where each utterance is treated independently.
To overcome the error propagation and further improve understanding performance, the contextual information has been leveraged and shown useful~\cite{bhargava2013easy,xu2014contextual,chen2015leveraging,sun2016an}.
Prior work incorporating dialogue contexts into the recurrent neural networks (RNN) for improving domain classification, intent prediction, and slot filling~\cite{xu2014contextual,shi2015contextual,weston2015memory,chen2016end}.
Recently, modeling speaker role information~\cite{chi2017speaker,chen2017dynamic,zhang2018addressee} has been demonstrated to learn the notable variance in speaking habits during conversations for better understanding performance.

% Figure
\begin{figure*}[t]
\centering
\includegraphics[width=\linewidth]{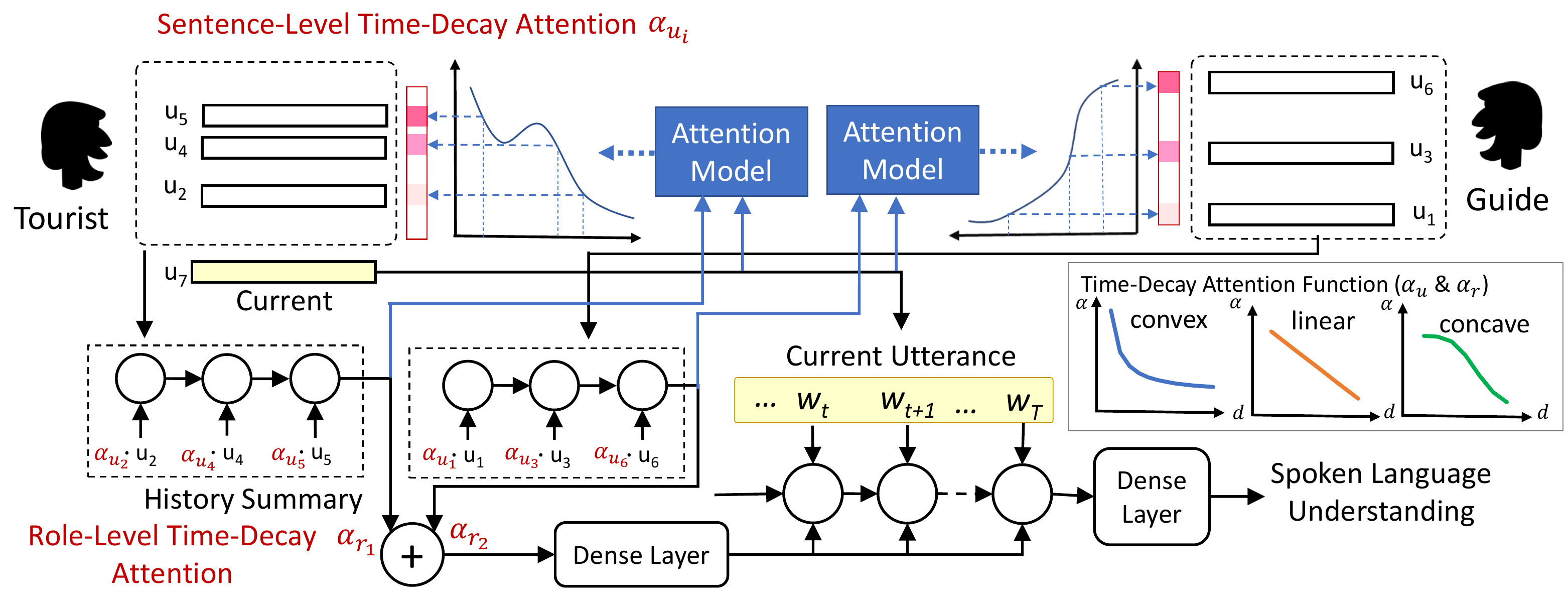}
\vspace{-2mm}
\caption{Illustration of the proposed role-based context-sensitive time-decay attention contextual model.}
\label{fig:model}
\vspace{-4mm}
\end{figure*}

Neural models incorporating attention mechanisms have advanced various tasks such as machine translation~\cite{bahdanau2014neural}, image captioning~\cite{xu2015show}, etc.
Attentional models have been successful because they separate two different concerns: 1) deciding which input contexts are most relevant to the output and 2) predicting an output given the most relevant inputs.
% For example, the highlighted current utterance from the tourist, ``\textit{uh on august}'', in the conversation of Figure~\ref{fig:example} is to respond the question about \texttt{WHEN}, and the content-aware contexts that can help current understanding are the first two utterances from the guide ``\textit{and you were saying that you wanted to come to singapore}'' and ``\textit{un maybe can i have a little bit more details like uh when will you be coming}''.
% Although content-aware contexts may help understanding, the most recent contexts may be more important than others. In the same example, the second utterance is more related to the \textit{when} question, so the temporal information can provide additional cues for the attention design. 
In dialogues, although content-aware contexts may help understanding~\cite{weston2015memory,chen2016end}, the most recent contexts may be more important than others, so the temporal information can provide additional cues for the attention design.
Prior work proposed an end-to-end time-aware attention network to leverage both contextual and temporal information for spoken language understanding and achieved the significant improvement, showing that the temporal attention can guide the attention effectively~\cite{chen2017dynamic,su2018how}.
However, the time-aware attention function is an inflexible, which is a fixed function of time for assessing the attention weights.
%which is simply defined as the reciprocal of time difference between the target utterance and the preceding sentence.

This paper focuses on learning a flexible time-aware attention mechanism in neural models, where the attention can be dynamically decided based on the contexts for better language understanding.
This work is built on top of the role-based contextual model by modeling role-specific contexts differently to design the associated time-aware attention functions for improving system performance.
The contributions are three-fold:
\begin{compactitem}
\item The proposed end-to-end learnable attention has great flexibility of modeling temporal information for diverse dialogue contexts.
\item This work investigates speaker role modeling in attention mechanisms and provides guidance for the future research about designing attention functions in dialogue modeling.
%\item This paper analyzes various aspects of time-decay attention mechanisms and provides guidance for the future research about designing attention functions in dialogue modeling.
\item The proposed model achieves the state-of-the-art understanding performance in the dialogue benchmark dataset.
\end{compactitem}
%In order to comprehend what tourist is talking about and imitate how guide reacts to these meanings, this work built on top of a role-based contextual model by modeling role-specific contexts differently for improving the system performance and further design and analyze associated time-aware attention functions.
%Furthermore, we propose an end-to-end learnable universal time-decay mechanism which demonstrates the flexibility to model the temporal information of diverse dialogue context and achieves a significant progress in performance of SLU in dialogues.

% This paper focuses on investigating various attention mechanism in neural models with contextual information and speaker role modeling for language understanding. 
% In order to comprehend what tourist is talking about and imitate how guide react to these meanings, this work proposes a role-based contextual model by modeling role-specific contexts differently for improving the system performance and further design associated time-aware and content-aware attention mechanisms.

\section{End-to-End SLU Framework}
\label{sec:framework}
The model architecture is illustrated in Figure~\ref{fig:model}.
First, the previous utterances are fed into the contextual model to encode into the history summary, and then the summary vector and the current utterance are integrated for helping understanding.
The contextual model leverages the attention mechanisms highlighted in red, which implements different attention functions for sentence and speaker levels.
%four attention mechanisms including 1) content-aware attention, 2) time-aware attention, 3) sentence-level attention and 4) role-level attention.
The whole model is trained in an end-to-end fashion, where the history summary vector and the attention functions are automatically learned based on the downstream SLU task.
%The objective of the proposed model is to optimize the conditional probability of intents given the current utterance, $p(\mathbf{y}\mid \mathbf{x})$, by minimizing the cross-entropy loss between the prediction and the target:
The objective of the proposed model is to optimize the conditional probability of the intents given the current utterance, $p(\mathbf{y}\mid \mathbf{x})$, by minimizing the cross-entropy loss between prediction and target $q(\mathbf{y}\mid \mathbf{x})$:
%\begin{comment}
\begin{equation}
\mathcal{L}=-\sum_{k}\sum_{z}q(y_k=z\mid \mathbf{x}) \log p(y_k=z\mid \mathbf{x}),
\end{equation}
where the labels $\textbf{y}$ are the labeled intent tags for understanding.
%\end{comment}

\subsection{Attentional Speaker-Aware Contextual SLU}
\label{ssec:clu}
% undefined
% \paragraph{Language Understanding (SLU)}
Given the current utterance $\textbf{x}=\{w_t\}^T_1$, the goal is to predict the user intents of $\textbf{x}$, including speech acts and associated attributes.
%, such as \texttt{QST\_WHERE}.
We apply the bidirectional long short-term memory (BLSTM) model~\cite{schuster1997bidirectional} to context encoding to learn the probability distribution of user intents.
\begin{eqnarray}
\label{eq:basic}
\textbf{v}_\text{o} &=& \text{BLSTM}(\textbf{x}, W_\text{his}\cdot \textbf{v}_\text{his}),\\
\textbf{o} &=& \mathtt{sigmoid}(W_\text{SLU}\cdot \textbf{v}_\text{o}),
\end{eqnarray}
where $W_\text{his}$ and $W_\text{SLU}$ are weight matrices and $\textbf{v}_\text{his}$ is the history summary vector. $\textbf{v}_\text{o}$ is the context-aware vector of the current utterance encoded by the BLSTM, and $\textbf{o}$ is the intent distribution.
Note that this is a multi-label and multi-class classification, so the $\mathtt{sigmoid}$ function is employed for modeling the distribution after a linear layer.
The user intent labels are decided based on whether the value is higher than a threshold tuned by the development set.

Considering that speaker role information is shown to be useful for better understanding in complex dialogues~\cite{chi2017speaker,zhang2018addressee},
we utilize the contexts from two roles to learn role-specific history summary representations, $\textbf{v}_\text{his}$ in (\ref{eq:basic}).
Each role-dependent recurrent unit $\text{BLSTM}_{\text{role}}$ receives corresponding inputs, $x_{t,\text{role}}$, which includes multiple utterances $u_i$ ($i=[1, ..., t-1]$) preceding the current utterance $u_t$ from the specific role, and have been processed by an encoder model.

There are various tasks showing the effectiveness of attention mechanisms~\cite{xiong2016dynamic,chen2016end}.
Recent work showed that two attention types (content-aware and time-aware) and two attention levels (sentence-level and role-level) significantly improve the understanding performance for complex dialogues.
This paper focuses on expanding the time-aware attention by learning dynamically context-sensitive time-decay functions in an end-to-end fashion.
For time-aware attention mechanisms, we apply it using two levels, sentence-level and role-level structures.
%, and Section~\ref{sec:time-decay} details the design and analysis of time-aware attention.

For the sentence-level attention, before feeding into the contextual module, each history vector is weighted by its time-aware attention $\alpha_{\text{role}_i}$:
\begin{eqnarray}
\textbf{v}^U_\text{his}=\sum_{\text{role}} \text{BLSTM}_{\text{role}}(x_{t,\text{role}}, \{ \alpha_{u_j} \mid u_j \in \text{role}\}),
\end{eqnarray}
\begin{comment}
\begin{equation}
\label{eq:tag2}
\textbf{v}_\text{his} =
\sum_{\text{role}}\textbf{v}_{\text{his}, \text{role}}
= \sum_{\text{role}}\text{BLSTM}_{\text{role}}(x_{t,\text{role}}),
\end{equation}
\end{comment}
where $x_{t, \text{role}}$ are vectors after one-hot encoding that represent the annotated intent and the attribute features.
Note that this model requires the ground truth annotations for history utterances for training and testing.
Therefore, each role-based contextual module focuses on modeling role-dependent goals and speaking style, and $\textbf{v}_\text{o}$ from (\ref{eq:basic}) would contain role-based contextual information.

\subsection{Universal Time-Decay Attention}
\label{sec:time-decay}

Because we assume that the most recent contexts are more important in dialogues, a time-aware attention should be a decaying function.
Considering that the contextual patterns may be diverse, a flexible and universal time-decay attention function that composes three types of attentional curves is formulated~\cite{su2018how}:
\begin{align}
\alpha^{\text{univ}}_{u_i} &= w_1 \cdot \alpha^{\text{conv}}_{u_i} + w_2 \cdot \alpha^{\text{lin}}_{u_i} + w_3 \cdot \alpha^{\text{conc}}_{u_i}\\
&= \frac{w_1}{a \cdot d(u_i)^b}+
w_2 (e \cdot d(u_i)+f) + \frac{w_3}{1+(\frac{d(u_i)}{D_0})^n}, \nonumber
\end{align}
where $w_i$ are the weights of time-decay attention functions, including three types~\cite{su2018how}: \emph{convex}, \emph{linear}, and \emph{concave}, illustrated in the top-right part of Figure~\ref{fig:model}.
Note that all attention weights will be normalized such that their summation is equal to $1$.

\begin{compactitem}
    \item \textbf{Convex} $\alpha^{\text{conv}}_{u}$:
Intuitively, recent utterances contain more salient information, and the salience decreases very quickly when the distance increases.
    \item \textbf{Linear} $\alpha^{\text{lin}}_{u}$:
The importance of preceding utterances linearly declines as the distance between the previous utterance and the target utterance becomes larger.
    \item \textbf{Concave} $\alpha^{\text{conc}}_{u}$:
Intuitively, the attention weight decreases relatively slow when the distance increases.
\end{compactitem}
Each of three types of decaying curves represents a different perspective on dialogue contexts and models different contextual patterns following the design in the prior work~\cite{su2018how}.

Because the framework can be trained in an end-to-end manner, all parameters ($w_i$, $a$, $b$, $e$, $f$, $D_0$, $n$) can be automatically learned to construct a flexible time-decay function.
With the combination of different curves and the adjustable weights, the model can automatically learn a properly oscillating curve in order to model the diverse and complex contextual patterns using the attention mechanism.

\subsection{Dynamically Context-Sensitive Attention}
\label{ssec:cstda}
As described in the previous sections, the proposed time-decay attention mechanisms have parameters ($a$, $b$, $e$, $f$, $D_0$, $n$) to determine the shapes of curves.
In addition to the time-decaying property, we further improve our design to dynamically encode context-sensitive characteristics into the associated attention weights. 
The feature vector $\textbf{v}_{\text{cur}}$ of the current utterances $\textbf{x}$ can be extracted by BLSTM or use the mean vector among pre-trained word embeddings of the current utterance. 
%\begin{equation}
%\textbf{v}_{\text{cur}} = \text{BLSTM}_{\text{cur}}(\textbf{x})
%\end{equation}

Considering that different speakers may have totally different speaking behaviors ~\cite{chi2017speaker,chen2017dynamic,zhang2018addressee}, a role-based context-sensitive attention is proposed. 
To better model the attention curve, the contextual information is also encoded by the $\text{BLSTM}$ model, where the preceding utterances from different speakers are encoded by different modules. 
\begin{align}
\textbf{v}_{\text{his},\text{role}} &= \text{BLSTM}_{\text{role}}(x_{t,\text{role}}), \\
\textbf{p}_{\text{role}} &= W_{\text{p},\text{role}}\cdot (\textbf{v}_{\text{his},\text{role}}, \textbf{v}_{\text{cur}}) + bias,
\end{align}
where the speaker-specific contextual encoding $\textbf{v}_{\text{his},\text{role}}$ is fed along with the feature of the current utterance ($\textbf{v}_{\text{cur}}$) into fully-connected layers to predict the parameters $\textbf{p}_{\text{role}} \in \{ a, b, e, f, D_0, n \mid \text{role} \}$ to determine the tendency of the attention curve.
Because the parameters $\textbf{p}_{\text{role}}$ are determined by the output of neural attention models without any clipping or projection and some of these uncontrolled real number are exponents, therefore the following two regularization terms are introduced as soft constraints, 
\begin{equation}
- \alpha \cdot \text{min}(\textbf{p}_{\text{role}}, 0) + \beta \cdot \sum \textbf{p}_{\text{role}}^2.  
\end{equation}
The first loss term is to encourage the model to output a positive number, and the second term is to facilitate the model to predict numbers with small absolute values, where $\alpha$ and $\beta$ are the weights to adjust the intensity of regularization.
Note that not all attention models use both regularization terms, while we endow the models with maximum flexibility and add constraints only if necessary.
For example, if the cut-off distance $D_0$ of the concave time-decay attention is negative, the denominator $1+(d(u_i)/{D_0})^n$ would easily become complex number, which is not applicable. 
To make $D_0\geq 0$, we use the model output as the exponent of the exponential function with $e$ as the base.
In order to further facilitate the concave decaying manner, the first term is applied; on the other hand, to prevent explosion, the second regularization term is utilized.

\begin{comment}
\subsection{End-to-End Training}
The objective is to optimize SLU performance, predicting multiple speech acts and attributes. % described in Section~\ref{ssec:clu}.
In the proposed model, all encoders, prediction models, and attention models can be automatically learned in an end-to-end manner.
\end{comment}

\section{Experiments}
\label{sec:experiments}
To evaluate the proposed model, we conduct the language understanding experiments on human-human conversational data.

\begin{comment}

\begin{table}
\centering
\begin{tabular}{ | l | c c c| }
    \hline
    \multicolumn{1}{|c|}{\multirow{2}{*}{\bf SLU Model}} & \multicolumn{3}{c|}{\bf Context Length}\\
 & 3 & 5 & 7\\
\hline \hline
No Attention Context & 74.75 & 74.69 (-) & 74.52 (-)\\
Content-Aware Context  & 74.04 & 73.90 (-) & 73.69 (-)\\
 \hline
Time-Aware (Hand)  & 76.05 & 76.34 (+) & 76.41 (+)\\
Time-Aware (E2E) & 76.26 & 76.43 (+) & 76.67 (+)\\
Content+Time (Hand) & 75.16 & 75.27 (+) & 75.48 (+)\\
Content+Time (E2E) & 75.82 & 75.92 (+) & 75.83 (-)\\
    \hline
    \bf Context-Sensitive & 76.62 & 76.96 (+) & 77.05 (+)\\
    \hline
  \end{tabular}
  \vspace{-2mm}
\caption{The sentence-level performance reported on F1 of the proposed time-decay attention under different context length settings (\%). `+' and `-' indicate the performance trends.}
\vspace{-2mm}
\label{tab:len}
\end{table}
\end{comment}

\begin{table*}
\centering
  \begin{tabular}{| c|  l l | c  c || c c c |}
    \hline
    \multicolumn{3}{|c|}{\bf SLU Model} & \bf Sentence-Level & \bf Role-Level  & \multicolumn{3}{|c|}{\bf Context Length}\\
    \multicolumn{3}{|c|}{} & &  & 3 & 5 & 7\\
    \hline\hline
     (a) & \multicolumn{2}{l|}{Na\"{i}ve SLU} & \multicolumn{2}{c||}{70.18} & \multicolumn{3}{c|}{--}\\
    \hline
    (b) & \multicolumn{2}{l|}{No Attention Contextual Model} & \multicolumn{2}{c||}{74.52} & 74.75 & 74.69 (-) & 74.52 (-)\\
    \hline
	(c) & \multicolumn{2}{l|}{Content-Aware Contextual Model~\cite{chi2017speaker}} & 73.69 & 74.28 & 74.04 & 73.90 (-) & 73.69 (-)\\
    \hline
   	(d) & Time-Decay Attentional Model~\cite{su2018how} &  Hand  & 76.41$^\dag$ & 76.68$^\dag$ & 76.05 & 76.34 (+) & 76.41 (+)\\
	(e) & &  E2E  & 76.67$^\dag$ &  76.75$^\dag$ & 76.26 & 76.43 (+) & 76.67 (+)\\
    \hline
    (f) & Content-Aware + Time-Decay Attention \cite{su2018how}&  Hand &  75.48$^\dag$ & 76.61$^\dag$ & 75.16 & 75.27 (+) & 75.48 (+)\\
	(g) &  &  E2E  & 75.83$^\dag$ & 76.74$^\dag$ & 75.82 & 75.92 (+) & 75.83 (-)\\
    \hline  
    (h) & \multicolumn{2}{l|}{\bf Context-Sensitive Time-Decay Attention}  &  \bf 77.05$^\dag$ & \bf 76.87$^\dag$ & 76.62 & 76.96 (+) & 77.05 (+) \\
    \hline
  \end{tabular}
  \vspace{-2mm}
\caption{The understanding performance reported on F-measure in DSTC4, where the context length is 7 for each speaker (\%). $^\dag$ indicates the significant improvement compared to all baseline methods ($p<0.05$ on the one-tailed t-test). Hand: hand-crafted; E2E: end-to-end trainable.}
\label{tab:res}
\vspace{-2mm}
\end{table*}

\subsection{Setup}
\label{ssec:settings}
%\textbf{Dataset}: 
The experiments are conducted using the DSTC4 dataset, which consist of 35 dialogue sessions on touristic information for Singapore collected from Skype calls between 3 tour guides and 35 tourists, including 31,034 utterances and 273,580 words~\cite{kim2016fourth}. 
All recorded dialogues with the total length of 21 hours have been manually transcribed and annotated with speech acts and semantic labels at each turn level.
The speaker information (guide and tourist) is also provided.
The human-human dialogues contain rich and complex human behaviors and bring much difficulty to all tasks.
We randomly selected 28 dialogues as the training set, 5 dialogues as the testing set, and 2 dialogues as the validation set. 

We focus on predicting multiple labels including intents and attributes, so the evaluation metric is an average F1 score for balancing recall and precision in each utterance.
The experiments are shown in Table~\ref{tab:res}, where we report the average results over more than three runs for both tourists and guides.
In all experiments, we use mini-batch \textit{Adam} as the optimizer with the batch size of 32 examples.
The size of each hidden recurrent layer is 128 or 64; since the proposed approach uses additional attention models to predict parameters of decaying curves, to fairly verify the effectiveness of the proposed method, smaller hidden recurrent layers (size = 64) are utilized in the proposed model (row (h)) and bigger ones are conducted in others (rows (b)-(g)).
We use pre-trained 200-dimensional word embeddings $GloVe$~\cite{pennington2014glove}. We only apply 40 training epochs without any early stop approach. 

In the training process, we can assign the attention models random targets to incorporate the supervised loss during the first few epochs to accelerate training. 
This paper simply sets a integer target for the attention model at the very beginning. Note that experiments show that our attention model can be train from scratch in an end-to-end manner without any supervised signal and achieve the same performance.

\subsection{Effectiveness of Time-Decay Attention}
To evaluate the proposed time-decay attention, we compare the performance with the na\"{i}ve SLU model without any contextual information (row (a)), the contextual model without any attention mechanism (row (b)), and the one using the content-aware attention mechanism (row (c)), where the attention can be learned at sentence and role levels.
It is intuitive that the model without considering contexts (row (a)) performs much worse than the contextual ones for dialogue modeling.
The rows (d)-(h) utilized the time-decay attention; rows (d)-(e) use only the time-decay attention; rows (f)-(g) model both content-aware and time-decay attention mechanisms together, where content-aware attention is directly estimated by concatenation of each context and the current utterance by a NN module.
There are two settings for time-decay attention learning: 
1) {\bf Hand}: hand-crafted hyper-parameters (rows (d) and (f)) and 2) {\bf E2E}: end-to-end training for parameters (rows (e) and (g)).
In the hand-crafted setting, the hyper-parameters $a = 1, b = 1, e = -0.125, f = 1, D_0 = 5, n = 3$ are adopted, the parameters are chosen to examine the effectiveness of each type of decaying curve,
where we choose the parameters such that the effectiveness of each type of decaying manner could be properly investigated (the linear one will be located between the two curves). 
In the end-to-end setting, all parameters are learnable parameter initialized as the hyper-parameters described above and fine-tuned by end-to-end learning. 
The row (e) previously achieves the state-of-the-art performance~\cite{su2018how}.
Our proposed context-sensitive time-decay attention model is shown in the row (h).

Table~\ref{tab:res} shows that all models with the time-decay attention (row (d)-(g)) outperform the model without temporal modeling.
However, row (c) performs worse than the one without any attention mechanism (row (b)), and rows (f)-(g) are slightly worse than the ones with only time-decay attention (rows (d)-(e)), revealing that without a delicately-designed attention mechanism, it is not guaranteed that incorporating an additional content-aware attention would bring improvement.
% By introducing time-decay attention, the experimental results show that all models with role-level attention and the some with sentence-level attention obtain the improvement, where the universal design of the proposed context-sensitive attention model (row (h)) achieves the state-of-the-art performance.

\subsection{Analysis of Context-Sensitive Attention}
Prior work (rows (f) and (g)) integrated both content-aware and time-decay attention to demonstrate the capability of mitigating the negative effect by the coarse design of content-aware attention model, but leveraging both attention types ironically results in worse performance than using single time-decay attention (row (d)-(e))~\cite{su2018how}.
The reasons may be that: 1) the harmful impact of low-quality content-aware attention is overwhelming, 2) the interaction between two types of attention during learning is not cooperative enough. 
Even though the row (g) in Table~\ref{tab:res} learns both content- and time-aware attention functions, the time-decay attention curve is fixed after training; in other words, it is not content-responsive.
If a history sentence contains salient information, it would be weighted by a small attention value from the time-decay attention curve regarding the large time difference. 

Our proposed context-sensitive attention model effectively integrates time-aware and content-aware perspectives, where instead of training the content-aware and time-aware attention separately, we utilize contextual information to dynamically construct the time-decay attention curves.
% , so-called ``context-sensitive time-decay attention''. 
% Considering that the parameters used to determine the tendency of decaying curves are predicted by neural models, we set two fully-connected layers in the attention models
%, where the universal time-decay attention uses smaller size due to the number of attention models, so that the results can be fairly compared. 
The results show that proposed role-based context-sensitive attention model (row (h)) outperform all compared baselines,
% Furthermore, the context-sensitive time-decay attention achieves the best performance, 
yielding 9.7\% improvement over the Na\"{i}ve baseline (row (a)). 
%
% Note that although the additional attention models and recurrent units are conducted, we cut the hidden layers of the LSTM units into half-size to keep the same model size.
% The proposed context-sensitive time-decay attention mechanism flexibly composes three types of time-aware attention functions in different decaying tendencies, each of which reflects a specific perspectives on distribution over salient information in dialogue contexts.
% It shows great capability of modeling diverse dialogue patterns in the experiments and therefore empirically shows that the proposed method is a general design of time-decay attention. 
%The design of the time-decay attention is simple, general and easily-extensible, the weights and the parameters in each type of attention can be assigned via hyperparameters (row (d) and (f)), initialized by hyperparameters and fine-tuned by end-to-end training (row (e) and (g)), or learned by fully end-to-end learning (row (h)).
% To further analyze the combination of different time-decay attention functions, we inspect the converged values of the trainable parameters from the proposed time-decay attention models.
% In the experiments, the models automatically figure out that convex time-decay attention function should have a higher weight than others for both sentence-level or role-level models ($w_1 > w_2$ and $w_1 > w_3$).
% In other words, it reflects that the majority of salient information lies in few recent utterances and once again demonstrates our claim.
%
As mentioned above, one can control the level of flexibility in the time-decay attention at will,
%
% ; for example, the proposed context-sensitive time-decay attention possesses the best flexibility.
% The more freedom means less controllable, so integrating different time-decay perspectives does not guarantee to improve the efficacy of the model. 
it is possible that the combination may interfere attention model learning.
Surprisingly, experiments show that the universal models outperform the models with a single time-decay attention type, demonstrating the positive interaction between attention functions and efficacy of our design.
% ; furthermore, the most flexible model (row (h)) achieves the best results.

\subsection{Speaker Role in Attention Modeling}
\label{ssec:sent-role}
For role-level attention, Table~\ref{tab:res} shows that all results with various time-decay attention mechanisms are better than the one with only content-aware attention (row (c)).
Considering the benefit of considering speaker interactions~\cite{chi2017speaker,chen2017dynamic}, therefore instead of weighting each utterance by its sentence-level attention, our model computes a representative attention value for each speaker by using the most important, representative utterances among what the speaker said. 
Namely, for role-level attention, each speaker role is assigned an attention value to represent the importance from the conversational interactions.
By introducing role-level attention, the sentence-level attention weights can be smoothed to avoid inappropriate values and benefit language understanding.
% Surprisingly, even though learning sentence-level temporal attention is difficult, our proposed universal time-decay attention can achieve similar performance for sentence-level and role-level attention (76.67\% and 76.75\% from the row (e)), further demonstrating the strong adaptability of fitting diverse dialogue contexts and the capability of capturing salient information.
Surprisingly, even though learning sentence-level temporal attention is difficult, the proposed context-sensitive time-decay attention (row (h)) is the only one whose sentence-level results are better, further demonstrating the strong adaptability of fitting diverse dialogue contexts and the capability of capturing salient information.

The proposed methods are built on top of the role-base contextual framework, which utilizes separate modules to learn speaker-specific features to improve understanding.
However, the prior time-decay attention models (rows (d)-(g)) are speaker-independent, where different speakers share the same decaying attention curve.
To further investigate the effectiveness of the speaker role in attention modeling, we make the proposed context-sensitive attention speaker-dependent, so-called ``role-based context-sensitive attention''.
The result (row (h)) shows that role-based attention modeling is promising, of which the universal design performs best.
In sum, our attention model design not only elegantly combines content-aware and time-aware perspectives but effectively integrates the concept of speaker role modeling into attention mechanisms.

\subsection{Robustness to Context Lengths}
% It is intuitive that longer context abounds richer information; however, it may obstruct the attention learning and result in poor performance because more information should be modeled and accurate estimation is not trivial.
It is intuitive that longer context abounds richer information; however, it may obstruct attention learning and result in poor performance due to too much information for digesting and more noises for inaccurate estimation.
% Because when modeling dialogues, we have no idea about how many contexts are enough for better understanding, the robustness to varying context lengths is important for the contextual model design.
Because when modeling dialogues, we have no idea about how many contexts are enough for better understanding, the robustness to varying context lengths becomes an important issue for contextual SLU.
Here, we compare the results using different context lengths (3, 5, 7) for detailed analysis in Table~\ref{tab:res}, where the number is for each speaker.
The results show that: 
1) the models without attention and content-aware attention become slightly worse with increasing context lengths;
% 2) the universal time-decay attention models in the row (d) to (g) in the Table~\ref{tab:res} mostly achieve better performance when conducting longer contexts, the model leveraging content-aware and time-aware attention by end-to-end learning (row (g) in Table~\ref{tab:res}) outperforms the one under handcrafted setting (row (g) in Table~\ref{tab:res}) whereas it weakens as context lengths become longer, showing less robustness. 
2) the time-decay attention models from the rows (d)-(g) in the Table~\ref{tab:res} mostly achieve better performance when conducting longer contexts, where the model leveraging content-aware and time-aware attention by end-to-end learning outperforms the one under handcrafted setting whereas it weakens as context lengths become longer, showing less robustness to context lengths; 
3) the proposed context-sensitive method performs the best for all context length settings, demonstrating not only the \emph{flexibility} of adapting diverse contextual patterns but also the \emph{robustness} to varying context lengths.

\section{Conclusion}
% This paper designs a role-based context-sensitive time-decay attention functions based on an end-to-end contextual language understanding model, where different perspectives on dialogue contexts are analyzed and a flexible and universal time-decay attention mechanism is proposed.
This paper designs a role-based context-sensitive time-decay attention functions based on an end-to-end contextual language understanding model, where different perspectives on dialogue contexts are analyzed.
The experiments on a benchmark human-human dialogue dataset show that the understanding performance can be boosted by introducing the proposed attention mechanisms which elegantly integrate content-aware, time-ware, speaker-role perspectives.
% for guiding the model to focus on the salient contexts following a convex curve.
% Moreover, the proposed method is easily extensible to multi-party conversations and showing the potential of leveraging temporal information in NLP tasks of dialogues.
Furthermore, the proposed method is easily extensible to multi-party conversations and showing the potential of integrating temporal and contextual information in NLP tasks of dialogues.

\vfill\pagebreak

% References should be produced using the bibtex program from suitable
% BiBTeX files (here: strings, refs, manuals). The IEEEbib.bst bibliography
% style file from IEEE produces unsorted bibliography list.
% -------------------------------------------------------------------------
\bibliographystyle{IEEEbib}
\bibliography{strings,refs}

\end{document}